\documentclass{ecai}

\usepackage{amssymb}
\usepackage{amsmath}
\usepackage{amsthm}
\usepackage{amsfonts}
\usepackage[numbers,square]{natbib}
\usepackage{graphicx}
\usepackage{placeins}
\usepackage{array}
\usepackage{multirow}
\usepackage{multicol}
\usepackage{dcolumn}
\usepackage{hyperref}
\usepackage{setspace}
\usepackage{tabularx}
\usepackage{booktabs}
\usepackage{float}
\usepackage[table,xcdraw]{xcolor}
\usepackage{soul}
\usepackage[most]{tcolorbox}
\usepackage{grfext}

\theoremstyle{definition}
\newtheorem{definition}{Definition}[section]

\theoremstyle{remark}
\newtheorem{remark}{Remark}[section]
\theoremstyle{plain}

\numberwithin{equation}{section}

\DeclareMathOperator{\sgn}{sgn}

\usepackage{todonotes}

\newcommand{\dsnote}[1]{\todo[color=green!30, inline]{DS: #1}}
\newcommand{\todonote}[2]{\todo[color=green!30, inline, caption={}]{TODO (#1): #2}}

\newcommand\de{\mathrel{\bullet\mkern-2.5mu{\rightarrow}}}

\newcolumntype{R}{>{\raggedleft\arraybackslash}X}
\newcolumntype{Z}{>{\centering\arraybackslash}X}
\newcolumntype{W}{>{\centering\arraybackslash}m{.75in}}

\ecaisubmission

\begin{document}


\begin{frontmatter}

\title{Machine Learning and Statistical Approaches to Measuring Similarity of Political Parties}

\author[A]{\fnms{Daria}~\snm{Boratyn} \orcid{0000-0003-3299-7071}}
\author[A]{\fnms{Damian}~\snm{Brzyski} \orcid{0000-0002-6867-1877}}
\author[A,B]{\fnms{Beata}~\snm{Kosowska-Gąstoł} \orcid{0000-0003-3555-2828}}
\author[C,D]{\fnms{Jan}~\snm{Rybicki} \orcid{0000-0003-2504-9372}}
\author[A,E]{\fnms{Wojciech}~\snm{Słomczyński} \orcid{0000-0003-2388-8930}}
\author[A,B]{\fnms{Dariusz}~\snm{Stolicki} \orcid{0000-0002-8295-0848}}

\thanks{This research has been funded under the Jagiellonian University Excellence Initiative, DigiWorld Priority Research Area, minigrant no. U1U/P06/NO/02.21 and QuantPol Center flagship project. Corresponding author: dariusz.stolicki@uj.edu.pl}

\begin{abstract}
Mapping political party systems to metric policy spaces is one of the major methodological problems in political science. At present, in most political science projects this task is performed by domain experts relying on purely qualitative assessments, with all the attendant problems of subjectivity and labor intensiveness. We consider how advances in natural language processing, including large transformer-based language models, can be applied to solve that issue. We apply a number of texts similarity measures to party political programs, analyze how they correlate with each other, and -- in the absence of a satisfactory benchmark -- evaluate them against other measures, including those based on expert surveys, voting records, electoral patterns, and candidate networks. Finally, we consider the prospects of relying on those methods to correct, supplement, and eventually replace expert judgments.
\end{abstract}

\end{frontmatter}


\section{Preliminaries}

Spatial models of politics, positing the existence of some metric (usually Euclidean) \emph{policy space} that bijectively maps to the universe of possible sets of political views and preferences, are central to many theoretical and empirical models of political behavior of voters, legislators, parties, and other political actors. For instance, such models can be used to explain election outcomes: we can assume that each candidate and each voter is positioned somewhere in that policy space, and that voters prefer candidates closer to their own ideal positions \citep{DavisEtAl70,EnelowHinich84}. Similarly, spatial models can be used to model party competition (parties seek positions that would attract most voters) \citep{AdamsEtAl05}, legislative decisionmaking (legislators vote for alternatives that are closer to their ideal points than the status quo) \citep{Tsebelis95,Tsebelis02}, or coalition formation (coalitions partners seek to minimize the distance of the expected coalition position and their own ideal point) \citep{Baron91,LaverBudge92,deVries99a,RusinowskaEtAl05}. A particularly simple example of a policy space is one-dimensional ordered metric space \cite{Black58,Downs57} which, in political science, is usually associated with the traditional left-right spectrum.

Clearly, to evaluate and apply spatial models it is essential to estimate the positions of the actors involved, and political parties are among the most important here 
\citep[p.~7]{Laver01}. The prevalent approach to this problem is still based on more or less structured, but ultimately qualitative human assessment, whether in the form of expert surveys, such as the Chapel Hill Expert Survey \citep{HoogheEtAl10,BakkerEtAl15,PolkEtAl17,JollyEtAl22}, V-DEM expert survey \citep{CoppedgeEtAl21,PemsteinEtAl21}, Global Party Survey \citep{Norris20}, or others \citep{HarmelEtAl95,Ray99,BenoitLaver06,RohrschneiderWhitefield06,RohrschneiderWhitefield16}, or of human coding of party political programs and electoral manifestos (see, e.g., \citealp{BudgeEtAl01,DolezalEtAl16,KlingemannEtAl06,VolkensEtAl18,LehmannEtAl23}). The former approach is subject to coder bias and subjectivity, with the resulting problems of reproducibility and reliability \citep{Budge00,SteenbergenMarks07}. There are also data availability issues: especially past party positions cannot be reliably coded 
\citep{Volkens07}. The document-based approach fares somewhat better in those respects (although it still involves subjective judgments) \citep{Keman07,McDonaldEtAl07,Volkens07}, but is more time-consuming \citep{EckerEtAl22}.

A natural solution would be to replace human coding with computerized content analysis \citep{PenningsKeman02}, and such attempts are already quite numerous. See, e.g., \citep{LaverEtAl03,BenoitLaver03,BenoitLaver06,KlemmensenEtAl07,Lowe08,SlapinProksch08,LauderdaleHerzog16}. Most of them follow the political science focus on programs and manifestos, being based on textual comparisons of such party documents. However, they have developed in relative isolation from rapid advances made in the field of natural language processing within the last 10 years. Thus, there have been strikingly few attempts to use methods like word embeddings or large language models for party positioning or similarity measurement, and no systematic comparisons or evaluations of their performance in this field. This is the gap we aim to fill.

Our main objective is threefold: we seek to review measures of party program similarity (both those already applied in this field and those used to evaluate textual similarity in other fields), analyze their correlation patterns, and evaluate their performance on real-life data from Poland (2001-2019). Because there is no single benchmark to use in that evaluation -- party positions and party similarity are not merely latent variables, but quite imprecisely defined and grasped by researchers -- we evaluate them against expert surveys, as well as a number of non-programmatic party similarity measures (voter behavior, candidate networks, and voting and coalition patterns).

\subsection{Contribution}

The principal contribution of this paper lies in \textbf{systematically testing, comparing, and benchmarking textual similarity measures and algorithms} developed in natural language processing and stylometry as applied to the similarity analysis of \textbf{political party programs}. As far as we are aware, for most of the said methods this is a pioneering application in this field. We also experiment with different \emph{hyperparameter choices} and \emph{document length normalization methods} designed to correct for differences in input lengths. Finally, in light of conceptual difficulties in defining party similarity, we \textbf{introduce and develop several benchmark measures} (coalition, genealogical, and electoral similarity indices are first introduced here).

\subsection{Prior Work}

Early research on the use of natural language processing for party positioning has been dominated by a \emph{single-minded focus on topic modeling} \citep{LaverGarry00,Ray01,KleinnijenhuisPennings01,LaverEtAl03,BudgePennings07,Pennings11,GrimmerStewart13}, at least initially mostly dictionary-based \citep{deVries99,Garry01,deVriesEtAl01}. Later, the standard toolbox of computerized party program analysis has been augmented with two ideological scaling algorithms -- WordScore and WordFish.

\emph{WordScore} is a supervised scaling / classification method developed in 2003 by Benoit, Laver, and Lowe \citep{LaverEtAl03,BenoitLaver03,BenoitLaver06,KlemmensenEtAl07,Lowe08}, and resembling a naive Bayes classifier. It calls for estimating, for every (non-stop) word in the corpus, a \emph{score vector}, i.e., a vector of probabilities that a given word appears in connection with a given label in the training set. For prediction, we average word score vectors over an input text. The result can be used either for scaling (with each coordinate corresponding to one dimension of the scaling space) or for classification (with the label corresponding to the largest coordinate being the predicted one).

\emph{WordFish}, developed by Slapin and Prokosch in 2008 \cite{SlapinProksch08}, is a term frequency-based method for unsupervised single-dimensional scaling. It is based on an assumption that word frequencies follow a Poisson distribution with the rate parameter depending on (latent) party position. Both the latent variables and coefficients are estimated using the expectation maximization algorithm. This method has been used in \citep{LauderdaleHerzog16} for ideological scaling of legislative speeches as well.

There exists voluminous literature on agreement between expert surveys and document coding (usually combined with some scaling method) \citep{BenoitLaver07,Keman07,McDonaldEtAl07}, as well as on comparing the two with other sources of data, usually voter and party elite surveys, party self-placements, or -- more recently -- voting advice application data \citep{MarksEtAl07,Ray07,Busch16,Lesschaeve17,EckerEtAl22,FerreiradaSilvaEtAl23}. Relatively few such studies incorporate behavioral data such as roll call voting records \citep{KrouwelvanElfrinkhof14} or coalition formation patterns \citep{Molder17}. Finally, some of the most recent works evaluate computerized content analysis methods, but none of them go beyond WordScores and WordFish \citep{HjorthEtAl15,BruinsmaGemenis19,RuedinMorales19}. Researchers tend to find high levels of agreement between expert surveys and other data sources, except that manifesto data diverge (although that may be the result of imperfect scaling rather than inherent problems) \citep{EckerEtAl22}.

\section{Textual Similarity Measures (text / styl)}

\subsection{Word Frequency Distributions}

The earliest algorithmic approach to textual similarity is to represent documents as (unordered) collections of words, informally referred to as \emph{bags-of-words} \citep[p.~19]{Eisenstein19}. Obviously, such a representation involves loss of contextual information carried by the segmentation of text into paragraphs and sentences and, more importantly, by their grammatical structure arising from word orderings. However, it is frequently employed for the sake of simplicity, and research suggests it exhibits relatively good performance, see, e.g., \citep{ShahmirzadiEtAl19}.

Without further loss of information we can map any bag-of-words to a probability distribution over individual words, thereby reducing the problem of measuring their similarity to a well-known problem of measuring similarity of discrete probability distributions \citep[ch.~14]{DezaDeza14}. Thus, the final representation of our corpus is an $n \times |V|$ matrix, where $n$ is the number of documents (party programs) and $V$ is the set of all words appearing in the corpus.

While researchers employing bag-of-words methods and word frequency distributions differ in what \emph{text preprocessing} techniques they apply before mapping the input text to a bag-of-words \citep{Chai22}, we opt for more extensive preprocessing in the form of \emph{case-normalization} (i.e., lowercasing), \emph{lemmatization}, and \emph{stop words removal}. This is because party programs are usually available only in their native form, i.e., in original languages, many of which are inflected languages, and because they are relatively short. Hence, in the absence of lemmatization, entropy of the word frequency would be artificially inflated, potentially distorting the results.

We experiment with several variants of word frequency-based methods: two \emph{word weighting methods} and three \emph{metrics}. The idea underlying word weighting is that variance in word frequency across documents increases with the expectation, wherefore more prevalent words have a much greater effect on the results than those less prevalent but more distinctive to the corpus or to individual documents \citep{SaltonBuckley88}. The standard correction for this, originating in the field of information retrieval but commonly used across all NLP fields, is the \emph{term frequency--inverse document frequency} (TF-IDF) measure, where each word is assigned a weight decreasing with probability that it occurs at least once within a random document in the corpus, and the weighted word vectors are normalized in such manner that their $L_2$ norms equal $1$ \citep{Aizawa03}. For our experiments, we test both unweighted word frequencies (\textbf{TF}) and \textbf{TF-IDF}.

With respect to \emph{the choice of a similarity measure}, we experiment with a number of standard functions. We denote the frequency matrix by $\mathbf{W}$, and its $i$-th column by $\mathbf{W}_i$.
\begin{description}
    \item[$L_1$ (Manhattan) metric] $d_{L_1}(i, j) := \lVert \mathbf{W}_i - \mathbf{W}_j \rVert_{1}$, which for stochastic vectors is identical up to a multiplicative constant to the total variation distance $d_{TV}$ between corresponding probability distributions;
    \item[$L_2$ (Euclidean) metric] $d_{L_2}(i, j) := \lVert \mathbf{W}_i - \mathbf{W}_j \rVert_{2}$;
    \item[cosine similarity] $s_{\cos}(i, j) := (\mathbf{W}_i \cdot \mathbf{W}_j) \big/ \left(\lVert\mathbf{W}_i\rVert_{2} \lVert\mathbf{W}_j\rVert_{2}\right)$.
\end{description}

\subsection{Stylometry}

Stylometry is usually regarded as use of statistical analysis of a text aimed at identifying its authorship by discerning author-specific style \cite{MostellerWallace63}. However, the body of scholarship on stylometric analysis of literary texts convincingly demonstrates that stylometry can also be applied to identify variables going beyond authorship, such as genre \cite{Douglas92,StamatatosEtAl00,Underwood16}, chronology \cite{Lutoslawski97,Hoover07,Rybicki21}, or overall sentiment \cite{StoneEtAl66,AcerbiEtAl13,Mohammad11,MohammadTurney10}. Accordingly, it is interesting to test whether party ideology can also be discerned through stylometric analysis.

The usual approach in stylometry is to compare frequency distributions of $N$ words that are most frequently used in the given textual corpus. This emphasis on frequently used words, including parts of speech commonly regarded as stop words in other NLP fields (such as conjunctions and prepositions), is particularly characteristic.

Two basic parameters for a stylometric similarity measure are the choice of the number of most frequently used words $N$ and the choice of the metric. With regard to the former, we experiment with $N = 50$, $100$, and $200$, noting that $100$ is fairly common in stylometric analyses. With regard to the latter, we test eight metrics:
\begin{description}
    \item[\textbf{cosine distance} (\textbf{styl-cos})] $d_{\cos}(i, j) = 1 - s_{\cos}(i, j)$,
    \item[\textbf{Burrows' delta} (\textbf{styl-delta})] $d_{\Delta}(i, j) = \lVert z(\mathbf{W})_i - z(\mathbf{W})_j \rVert_{1}$, where $z: \mathbb{R}^{n \times |V|} \rightarrow \mathbb{R}^{n \times |V|}$ is a row-wise standardization \citep{Burrows02},
    \item[\textbf{Argamon's rotated quadratic delta} (\textbf{styl-arg}),] which differs from Burrows' delta in that $L_2$ rather than $L_1$ norm is used and the word frequency matrix is rotated using eigenvalue decomposition according to the word frequency covariance matrix calculated from the whole corpus \citep{Argamon08};
    \item[\textbf{Eder's delta} (\textbf{styl-eder}),] which differs from Burrows' delta by applying an inverse-frequency-rank weight to words \citep{EderEtAl16};
    \item[\textbf{cosine delta} (\textbf{styl-cosd}),] which differs from Burrows' delta in that cosine rather than $L_1$ distance is used \citep{JannidisEtAl15};
    \item[\textbf{cross-entropy} (\textbf{styl-entropy})] $d_{H}(i, j) = -\sum_{k=1}^{|V|} \mathbf{W}_{ik} \log \mathbf{W}_{jk}$ \citep{JuolaBaayen05};
    \item[\textbf{minmax distance} (\textbf{styl-minmax})] $\min\{\mathbf{W}_{i}, \mathbf{W}_{j}\} / \max\{\mathbf{W}_{i}, \mathbf{W}_{j}\}$, where $\min$ and $\max$ are taken element-wise \citep{KestemontEtAl16};
    \item[\textbf{Eder's simple distance} (\textbf{styl-simple})] $d_{L_1}(i, j) := \lVert \sqrt{\mathbf{W}_i} - \sqrt{\mathbf{W}_j} \rVert_{1}$, with the square-root taken element-wise \citep{EderEtAl16}.
\end{description}

\noindent For all stylometric computations, we use \texttt{stylo} package for R by Eder et al. \cite{EderEtAl16}.

\subsection{Static Word Embeddings}

Methods based purely on word frequency do not account for semantics. In essence, they correspond to the assumption that the semantic metric on the space of words is discrete, i.e., that all distinct words are equidistant. Clearly, this assumption is a substantial oversimplification. Accordingly, we also use methods that account for semantic rather than lexical similarity, starting with methods employing \emph{distributional word embeddings}. Such embeddings are injective functions mapping each word to an element of some finite-dimensional metric space $(\mathcal{E}, d)$ in such manner that distances in $(\mathcal{E}, d)$ decrease as the corresponding words become more semantically similar. Semantic similarity, in turn, is operationalized on the basis of co-occurrence statistics, invoking Firth's \emph{distributional hypothesis} \cite{Firth57}, according to which the more semantically similar two words are, the more likely they are to appear interchangeably in the same context.

We begin with \emph{static} embedding methods, characterized by being context-invariant, i.e., always representing identical words in the same manner. We focus on three arguably most common embedding algorithms: the original \emph{word2vec} algorithm by Mikolov et al. \cite{MikolovEtAl13} (see also \cite{MikolovEtAl13a,MikolovEtAl15}); \emph{FastText} algorithm by Bojanowski et al. \cite{BojanowskiEtAl17}, designed to account for morphological properties of the space of words; and \emph{GloVe} algorithm by Pennington et al. \cite{PenningtonEtAl14}. For all three, the codomain $\mathcal{E}$ is a high-dimensional linear space. We experiment with different values of $\dim \mathcal{E}$: $100$, $300$, $500$, and $800$ for word2vec and FastText, and $300$ and $800$ for GloVe. The choice of these values has been dictated by the availability of pretrained models.

How word embeddings can be used to compare party programs (or other texts, for that matter)? One standard approach is to map the whole text to vector representation word-by-word, and then aggregate by averaging over all words. Another one is to use \emph{word mover's distance} (WMD) \citep{KusnerEtAl15}, which is essentially the \emph{Wasserstein metric} over the embeddding codomain. Estimation of this distance is a common variant of the Kantorovich-Monge transportation problem \cite{FordFulkerson56}, which we solve using the \textit{displacement interpolation} algorithm \cite{BonneelEtAl11,FlamaryEtAl21}. We experiment with both approaches, yielding us a total of twenty methods per corpus.

\subsection{Transformer-Based Language Models}

More advanced language models incorporate contextual information when mapping words to their vector representation. While formerly such models were based on recurrent neural networks or long-short term memory (LSTM) networks \cite{Goldberg17}, since ca. 2018, \emph{transformers} (recursive self-attention-based networks) \citep{VaswaniEtAl17} have been widely regarded as the state-of-the-art solution \citep{WolfEtAl20}. Most transformer-based models are still word embedding models, although capable of utilizing contextual information from the input text. We experiment with four such models: GPT-2 \cite{RadfordEtAl18} (trained using a left-to-right encoder), RoBERTa \cite{LiuEtAl19} (trained using a bidirectional encoder), BART \citep{LewisEtAl19} (trained using a composite noising scheme), and LongFormer \citep{BeltagyEtAl20} (with linearly-scaling attention mechanism, which enables it to admit longer token sequences for context).

In addition, we consider large language models capable of mathematically representing variable-length chunks of texts (usually sentences) rather than individual words. These include Sentence-BERT \citep{ReimersGurevych19}, using two BERT \citep{DevlinEtAl18} encoders and a pooling model; Universal Sentence Encoder \citep{CerEtAl18}, combining a transformer base model with deep averaging network; and DefSent \citep{TsukagoshiEtAl21}, trained on definition sentences from dictionaries. In the present article we use SBERT-based \emph{Sentence Transformers} library \citep{Reimers22}, experimenting with models trained using MPNet \cite{SongEtAl20} and DistilRoBERTa \cite{SanhEtAl20,SanhEtAl21}.

For aggregation of the resulting representations, we use two methods that are most common in transformer-based models literature: averaging over words (sentences) (\emph{mean pooling}), and taking an elementwise maximum over the same (\emph{max pooling}).

\subsection{Methods for Dealing with Length Differences}

Significant differences in text length can be a major source of distortion in most if not all of the methods described in preceding subsections. However, such differences are common among political party programs. For example, in our reference dataset the ratio of the lengths (in characters) of the largest and shortest program equals approx. $246$, and the standard deviation of the natural logarithm of text length is $1.31$. While differences at a single point in time tend not to be that extreme (usually the max-min ratio on the order of $10$ to $30$), they are still sufficiently large to raise doubts about comparability. To assuage those doubts, we experiment with two methods for dealing with text length differences: \emph{random sampling} and \emph{summarization}.

We use two kinds of random sampling techniques. For stylometry, we sample individual words uniformly with replacement and average the results over $256$ samples (leveraging the sampling procedure built into the \texttt{stylo} package). For other methods, we divide the text into sentences using \texttt{spacy} library for Python \citep{HonnibalMontani23}, uniformly sample $120$ sentences, and average the results over $256$ samples.

For summarization we use the following algorithm. First, the text is divided into $25$ connected chunks in such manner that the difference in the number of sentences in the longest and shortest chunks is at most $1$. Second, from any chunk of more than $4$ sentences we choose exactly $4$ sentences whose vector representation under the TF-IDF transform are closest to the TF-IDF vector of the whole chunk. From shorter chunks, we simply choose all sentences. The summary is obtained by concatenating all chosen sentences in the order in which they appear in the text. The advantage of this procedure lies in the fact that every sentence in the summary appears in the original text. Generative summarization would threaten to introduce artifacts that could disrupt some of our methods.

\section{Benchmark Measures}

The fundamental problem in choosing a benchmark for party similarity measures lies in the fact that the very concept of party similarity -- or the dual concept of party position in policy space -- is quite fuzzy and only very imprecisely grasped by researchers. Hence, we have \emph{no objectively correct similarity measure} to benchmark against. Instead, we test program similarity measures against standard methods in the field (various expert surveys) as well as against similarity measures for other areas of party activity (legislative voting, coalition formation, candidate selection, electoral campaigning). By the assumptions of spatial models, all of those should be correlated with proximity in the policy space, and therefore also with program similarity.

The basic challenge here is that those assumptions might not be fully (or even at all) satisfied. Domain experts tend to recognize that party similarity is multidimensional. Moreover, it might very well be the case -- indeed, many if not most political scientists would agree that it is the case -- that party programs are only imperfect representations of party views of policy, the divergence being attributable to strategic considerations: parties may include issues and promises that are not intrinsically important to them, but respond to current concerns of the electorate, and may obscure their positions on other issues if they judge such positions to be liabilities. This is likewise true of all other dimensions of party similarity. As for expert opinions, divergence between expert judgments and textual analysis of political programs may occur because the former incorporate other dimensions as well, or because the former are biased or distorted by misperception of actual party objectives. Accordingly, extreme care is needed in interpretation of the benchmarks.

\subsection{Expert Surveys}

As noted in Sec.~1, expert surveys are the standard source of party positioning data in political science. They are for the most part semi-structured: experts are asked to position the party on some given ordinal or interval scale for a number of issues defined in advance by survey authors (for instance, the Chapel Hill Expert Survey asks experts to position parties according to their views on economic policy, social and cultural issues, European integration, immigration, environmental sustainability, civil liberties, deregulation, etc.). In general, experts are not given any further instructions on how to map specific party positions to points on the survey scale.

The principal advantage of expert surveys lies in the fact that they are holistic: experts can integrate all kinds of different data sources and have maximum flexibility in aggregating them \citep{LaverGarry00,McDonaldMendes01,BudgePennings07}. On the other hand, the major weakness are reliability concerns \citep{Mair01,SteenbergenMarks07,Volkens07,HoogheEtAl10}. The very flexibility and lack of precise constraining criteria makes expert assessments less comparable and therefore more difficult to aggregate \citep{CastlesMair84,Keman07}. There are also obvious risks of experts' biases and misperceptions.

\subsubsection{Chapel Hill Expert Survey}

The leading expert survey on party positions is the Chapel Hill Expert Survey \citep{HoogheEtAl10,BakkerEtAl15,PolkEtAl17,JollyEtAl22}, dating back to 1999.
The first survey was conducted in 1999 and only included 14 West European countries, but subsequent iterations in 2002, 2006, 2010, 2014, and 2019 quickly expanded its scope.
The latest survey in 2019 covered all 28 EU member states (including the UK), as well as several non-EU states. Between 1999 and 2019, the number of national parties included in the CHES dataset increased from 143 to 268. The experts assess party positions on general left-right ideological axis, economic left-right axis, and the progressive-conservative axis (GAL-TAN), as well as on more specific issues such as European integration, immigration, or environment.

We consider four benchmarks based on CHES:
\begin{description}
    \item[\textbf{lrgen}] absolute difference of the values of CHES \textbf{lrgen} variables, defined as `position of the party ... in terms of its overall ideological stance';
    \item[\textbf{lreco}] absolute difference of the values of CHES \textbf{lrecon} variables, defined as `position of the party ... in terms of its ideological stance on economic issues';
    \item[\textbf{galtan}] absolute difference of the values of CHES \textbf{galtan} variables, defined as `position of the party ... in terms of their views on social and cultural values';
    \item[\textbf{ch2d}] $L_2$ (Euclidean) distance of the points in a two-dimensional space defined by CHES \textbf{lrecon} and \textbf{galtan} variables.
\end{description}

\subsubsection{V-DEM}

V-DEM (Varieties of Democracy Project) is a large comparative expert survey of different aspects of the functioning of democracy V-DEM expert survey \citep{CoppedgeEtAl21,PemsteinEtAl21}. One of its component parts is V-PARTY, a survey of parties and party systems, containing data on 3467 parties from 178 countries, in some cases dating back as early as 1900. We consider one benchmark based on V-DEM:
\begin{description}
    \item[\textbf{vdem}] $L_2$ (Euclidean) distance between vectors of V-PARTY ideological variables.
\end{description}

\subsubsection{Global Party Survey}

Global Party is one of the newer major party surveys, initiated by Norris \citep{Norris20}. It includes data about 1043 parties from 163 countries. We do not use Global Party Survey as a benchmark, because data from this source is only available for 2018.

\subsection{Manifesto Research Project (\textbf{MARPOR})}

The leading document-based party positioning effort is the Manifesto Research Project, currently financed by a long-term funding grant from the German Science Foundation (DFG) as Manifesto Research on Political Representation \citep{LehmannEtAl23}.
It continues the work of the Manifesto Research Group (MRG 1979-1989) and the Comparative Manifestos Project (CMP 1989-2009). The project has generated a data set based on the content analysis of electoral manifestos of the major political parties in mainly the OECD and CEE countries.
It covers over 1000 parties from 1945 until present in over 50 countries on five continents. To create the data set, trained native-language experts are asked to divide the electoral programs into statements (sentences or quasi-sentences, each containing a certain idea or meaning) and to allocate these quasi-sentences into a set of policy categories. This coding scheme comprises 56 categories that are divided into seven domains. The coding outcome is a single topic distribution vector for each manifesto. The theoretical basis of the MARPOR approach lies within the salience theory that understands competition among parties in terms of the distinct emphases the parties place on certain policy areas \cite{Budge15}. Quality and reliability of MARPOR data has been assessed by numerous scholars as relatively good, albeit subject to certain caveats \citep{VolkensEtAl09,BenoitEtAl09,MikhaylovEtAl12,Gemenis13,MeyerJenny13}.

Most domain experts either believe that policy spaces are low-dimensional, or at least prefer to work with such spaces for the sake of simplicity and interpretability, and therefore find MARPOR's $56$-dimensional topic distribution vectors not fully satisfactory. Accordingly, there exists quite extensive literature on the subject of \emph{scaling} MARPOR data (or other topic distribution data). The initial Manifesto Research Group approach to this problem was to use factor analysis for scaling \citep{BudgeEtAl87}, but because of sampling adequacy problems caused by the number of variables exceeding the number of observations, as well as interpretability issues, the authors ultimately settled on a much simpler solution -- the \textbf{RILE} (right-left) indicator, defined as a linear combination of a subset of coordinates assigned either positive or negative unit coefficients \citep{LaverBudge92}. Initial values of coefficients were assigned according to the \emph{a priori} judgment of domain experts, but factor and correlation analysis was then used to refine those assignments (although in a computer-assisted rather than purely algorithmic manner) \citep{BudgeRobertson87,Klingemann87,BudgeKlingemann01}.

While commonly used (see, e.g., \citep{PenningsKeman94}), RILE has met with extensive criticism \citep{Pelizzo03,DinasGemenis10,Molder16}, and a number of alternatives have been proposed, ranging from nonlinear transforms of the RILE measure \citep{LoweEtAl11} and different coefficient assignment methods \citep{Prosser14,FranzmannKaiser06}, to more sophisticated statistical techniques such as principal factor analysis \citep{GabelHuber00,Jahn11}, factor analysis on Q-transformed dataset \citep{PappiShikano04}, structural equation modeling \citep{Bakker09}, and latent variable analysis \citep{Elff08,Elff13,KonigEtAl13,FlentjeEtAl17}. A significant barrier to adoption of the latter class of methods, however, lies in the fact that they learn dimensions of the policy space from the data rather than permit the researcher to specify them \citep{FranzmannKaiser06}.

We consider two benchmarks based on MARPOR:
\begin{description}
    \item[\textbf{marpor}] cosine similarity of MARPOR topic distribution vectors;
    \item[\textbf{rile}] absolute difference of the values of the \textbf{rile} variable.
\end{description}

\subsection{Voting Agreement (\textbf{vote-kappa})}

Applying a spatial model of politics to legislative decision-making, we can consider a parliamentary vote on a contested issue as equivalent to a bisection of the policy space. It follows that, on average, parties close to each other in that space should vote in agreement more frequently than those distant from each other. Conversely, agreement in voting patterns is likely to imply policy proximity. Accordingly, we can treat voting agreement as a possible benchmark for our program similarity measures.


To quantify voting agreement between two parties we make some general assumptions. Firstly, we assume that (roll-call) voting in the parliament is \textit{ternary} with the 'abstain' (\textit{A}) option located exactly halfway between 'no' (\textit{N}, nay) and 'yes' (\textit{Y}, yea) options. This leads to the symmetric three-by-three \textit{agreement matrix} quantifying the similarity of the votes of two MPs in a particular voting. The matrix has values $1$ on the diagonal, i.e., if the votes cast are identical, $1/2$ for pairs (\textit{A}/\textit{N}) or (\textit{A}/\textit{Y}), and $0$ if the votes are opposite (\textit{N}/\textit{Y}). Secondly, in a concrete voting we calculate the mean value of this agreement index for a pair of random voters from these two parties. Thirdly, we average such obtained index over all votes with weights proportional to the products of the turnout (participation) of both parties in a given vote. Finally, we use the technique invented by Cohen \citep{Cohen60,Cohen68} and modified later by Vanbelle \citep{Vanbelle09,VanbelleAlbert09} to exclude the possibility of agreement occurring just by chance between the results of votes of both parties, obtaining in this way so called \textit{modified $\kappa$ coefficient} \citep{SlomczynskiStolicki15a,SlomczynskiStolicki16}.

There remains one question: which votes should we consider in the calculation of the $\kappa$ coefficient? One option is to look backward from the point in time at which we compare parties, in essence assuming the perspective of a voter at an election, able only to assess the past track record. Another option is to look forward, assessing how the party is carrying out its declarations. Both approaches appear to us equally valid, so we aggregate them into one by averaging them.

\begin{remark}
    One obvious weakness of using voting records as a benchmark lies in natural incompleteness of the data: parties that are not represented in the legislature in a given term have no voting record. If only forward or backward data are missing, we omit the averaging and just take the other value. If both forward and backward data are missing, we omit this benchmark for a given party.
\end{remark}

\subsection{Coalition Patterns (\textbf{coal})}

The organizing principle of interparty interactions in most parliaments is the government-opposition divide: most votes divide parties into those supporting and those opposing the government and there exists a coalition of parties that consistently vote with the government. However, if we assume parties to be rational actors that maximize the proximity of voting outcomes to their policy position, it follows that coalitions should form between parties that are close to each other. Accordingly, from the coalition formation patterns we should be able to make inferences about party proximity.


The simplest conceivable measure of coalition-based similarity is a Boolean one that assumes $1$ if two parties are coalition partners and $0$ otherwise. But this measure fails to account for the fact that failure to form a coalition will frequently stem not from interparty distance but rather from the fact that such coalition would not command a majority. Thus, proximity of opposition parties would be consistently underrated. One possible solution is to treat both coalition partners and co-opposition parties in the same manner. 
However, this might in turn overrate similarity of the opposition parties: two parties may be in the opposition together not because they are close to one another, but because they are both distant from the governing parties. An enemy of one's enemy is not necessarily one's friend. To reflect this, the \emph{ternary measure} of coalitional similarity assigns $1$ to coalition partners, $1/2$ to parties that are together on the opposition side, and $0$ to parties that are on different sides. Since we compare parties at the time of a general election, we ascertain this value for every day of the preceding and succeeding parliamentary terms, calculate a \emph{day-by-day average} (assuming that more durable coalitions imply greater similarity) for each term, and average the two values together.

\subsection{Genealogical Similarity (\textbf{cand-gen})}

While political scientists frequently follow the constructivist paradigm, treating parties as at least quasi-unitary actors distinct from the collection of their members, it is rather difficult to imagine party position in the political space to be wholly independent from the positions of its members, and especially from the positions of the party elite. At the same time, while some party systems have been stable for generations, (see, e.g., United States, Australia, Switzerland, or Japan, and to a more limited extent Germany and United Kingdom),
others are in flux (France, Italy, Poland).
In the latter, many politicians moved through several parties in the course of their careers. If, on the average, politicians from the same party are closer to each other than those from different parties, we would expect parties whose members (or at least elites) come from the same past party to be more similar than those whose members do not have such a shared background. This concept of \emph{genealogical similarity} is used to define our next benchmark.

As a first step, we construct a directed \emph{genealogical graph} $\mathcal{G}$. Let a set of elections in a given jurisdiction be indexed by $L \in \mathbb{N}$, let $P(i)$ for $i \in L$ be the set of all parties contesting the $i$-th election, and let party be identified with the set of its candidates. For the purpose of this definition, we ignore party continuity, so even if a party $X$ contested an election $i$ and then an election $j$, we treat $X$ as of $i$ and $X$ as of $j$ as distinct entities. The set of vertices of $\mathcal{G}$ equals $\bigcup_{i \in L} P(i)$, and an edge exists from $p$ to $q$ if and only if $p \cap q \neq \emptyset$ and there exists such $i \in L$ that $q \in P(i)$ and $p \in P(i+1)$, i.e., the two parties have common candidates and contested consecutive elections. A vertex $x \in V(\mathcal{G})$ is an \emph{ancestor} of $y \in V(\mathcal{G})$ if and only if there exists a path in $\mathcal{G}$ from $x$ to $y$.

Each edge $p \de q$ in the genealogical graph is assigned a weight:
\begin{itemize}
    \item for countries using non-party-list electoral systems, the weight is equal to $|p \cap q| / |p|$, i.e., the proportion of candidates in $p$ that belonged to $q$,
    \item for countries using party-list electoral systems, the weight is equal to $w(p \cap q) / w(p)$, where $w$ is an additive measure on $p$ such that $w(\{c\}) = r_c^{-1}$, where $c \in p$ is a candidate and $r_c$ is that candidate's position on the party list. Intuitively, this is equivalent to the non-party-list case, except that candidates are weighted inversely to their position on the party list in the later election.
\end{itemize}
The \textit{weight} of a \textit{path} in $\mathcal{G}$ equals the product of edge weights. For any fixed parties $p, q$ we denote the shortest path from $p$ to $q$ that is maximal in terms of weight by by $\pi(p, q)$.

\begin{definition}[Genealogical Similarity Measure (\textbf{cand-gen})]
The \emph{genealogical similarity measure} of parties $x$ and $y$ is given by:
\begin{equation}
    G(x, y) = \sum_{z \in A(x, y)} \min \left\{ w(\pi(x, z)), w(\pi(y, z)) \right\},
\end{equation}
where $A(x, y)$ is the set of common ancestors of $x$ and $y$.
\end{definition}

\subsection{Electoral Similarity (\textbf{elec-cor})}

As applied to electoral behavior, the spatial model posits that electorates of two parties that are close to each other should be similar in terms of their positions in the policy space. Accordingly, we would expect the vote shares of two similar parties to be correlated. Hence, our final benchmark is the \emph{electoral similarity measure} which for any two parties equals the \emph{Pearson correlation coefficient} of their municipal-level vote shares as of the most recent national parliamentary election, with the correlation taken over all municipalities.

\section{Data}

Finding a good dataset for testing of party similarity measures is surprisingly difficult, as much of the needed data is only available in digital form for quite recent elections. While program texts are available from MARPOR \cite{MerzEtAl16,LehmannEtAl23a}, benchmark data are incomplete and scattered over multiple sources. Our ideal dataset should cover several electoral cycles and include multiple parties per each election. These conditions are satisfied by a dataset of Polish electoral and party database for the 2001-2019 period, which includes a collection of digitized program texts (originally from \cite{Slodkowska01,Slodkowska01a,Slodkowska02,SlodkowskaDolbakowska04,SlodkowskaDolbakowska06,SlodkowskaDolbakowska11,SlodkowskaDolbakowska13,Slodkowska18}), a candidate database with personally unique keys that allow us to track candidates between elections, a database of precinct-level election results, and a dataset of legislative roll call voting records.

The dataset consists of 41 party electoral programs, which gives us 820 distinct pairs of programs to compare. We calculate inter-measure correlations for all such pairs. However, because several of our benchmarks can only be defined for parties existing at the same moment in time (for instance, we cannot compare if a party from 2001 and a party from 2015 voted in the same manner, because they participated in different roll call votes), we compare our similarity measures with benchmarks only for such parties.

Pretrained word embeddings and language models for Polish texts have been obtained from \cite{DadasEtAl22}.

\section{Results and Analysis}

\subsection{Preliminary Test -- Self-Similarities}

As a preliminary test, for each method of similarity measurement we have run a self-similarity test. Every party program in our corpus that was at least $32,768$ characters long was divided into two parts, one consisting only of odd sentences, and the other consisting only of even sentences, and then the methods tested were used to compare those parts. 
The distribution of the results was then compared with the distribution of inter-party similarities.

\subsection{Intra-Group Correlations}

Within each group of text analysis methods (word frequency, stylometry, static word embeddings, transformer word embeddings, transformer sentence embeddings) we compute a correlation matrix, and then use \emph{hierarchical agglomerative clustering} \citep{MurtaghContreras12}, iteratively merging clusters that are most correlated. We use \emph{Pearson's correlation coefficient} for quantifying correlations between two singleton clusters; \emph{multiple correlation coefficient} for quantifying correlations between a singleton cluster and a non-singleton cluster \citep[\S~6.2.2]{CarrollGreen97}; and \emph{group correlation coefficient} for quantifying correlations between two non-singleton clusters \citep{FilzmoserHron09}. However, we do not merge clusters if a merger would cause the minimal intra-cluster correlation to fall below $.75$ threshold. Because most groups of variables are rather numerous, we only report cluster composition and correlations between clusters.

\subsubsection{Measures Based on Word Frequency Distributions}


\begin{center}
\begin{tabular}{cc|ccc|}
\cline{3-5}
& & \multicolumn{3}{c|}{\textbf{length correction}} \\ \hline
\multicolumn{1}{|c|}{\textbf{method}} & \textbf{metric} & \multicolumn{1}{c|}{none} & \multicolumn{1}{c|}{sampling} & summarization \\ \hline
\multicolumn{1}{|c|}{TF} & cos & \multicolumn{1}{c|}{\cellcolor[HTML]{EFEFEF}1} & \multicolumn{1}{c|}{\cellcolor[HTML]{9B9B9B}3} & \cellcolor[HTML]{9B9B9B}3 \\ \hline
\multicolumn{1}{|c|}{TF} & $L_2$ & \multicolumn{1}{c|}{\cellcolor[HTML]{EFEFEF}1} & \multicolumn{1}{c|}{\cellcolor[HTML]{9B9B9B}3} & \cellcolor[HTML]{9B9B9B}3 \\ \hline
\multicolumn{1}{|c|}{TF} & $L_1$ & \multicolumn{1}{c|}{\cellcolor[HTML]{C0C0C0}2} & \multicolumn{1}{c|}{\cellcolor[HTML]{656565}4} & \cellcolor[HTML]{656565}4 \\ \hline
\multicolumn{1}{|c|}{TFIDF} & cos & \multicolumn{1}{c|}{\cellcolor[HTML]{EFEFEF}1} & \multicolumn{1}{c|}{\cellcolor[HTML]{9B9B9B}3} & \cellcolor[HTML]{9B9B9B}3 \\ \hline
\multicolumn{1}{|c|}{TFIDF} & $L_2$ & \multicolumn{1}{c|}{\cellcolor[HTML]{EFEFEF}1} & \multicolumn{1}{c|}{\cellcolor[HTML]{9B9B9B}3} & \cellcolor[HTML]{9B9B9B}3 \\ \hline
\multicolumn{1}{|c|}{TFIDF} & $L_1$ & \multicolumn{1}{c|}{\cellcolor[HTML]{C0C0C0}2} & \multicolumn{1}{c|}{\cellcolor[HTML]{656565}4} & \cellcolor[HTML]{656565}4 \\ \hline
\end{tabular}
\end{center}

\begin{center}
\begin{tabular}{c|c|c|c|c|}
\cline{2-5}
 & \multicolumn{4}{|c|}{\textbf{Inter-cluster correlation matrix}}
\\ \cline{2-5}
& {\cellcolor[HTML]{EFEFEF}1} & {\cellcolor[HTML]{C0C0C0}2} & {\cellcolor[HTML]{9B9B9B}3} & {\cellcolor[HTML]{656565}4} \\ \hline
\multicolumn{1}{|c|}{\cellcolor[HTML]{EFEFEF}1} & 1.000 & 0.950 & 0.996 & 0.945 \\ \hline
\multicolumn{1}{|c|}{\cellcolor[HTML]{C0C0C0}2} & 0.950 & 1.000 & 0.908 & 0.901 \\ \hline
\multicolumn{1}{|c|}{\cellcolor[HTML]{9B9B9B}3} & 0.996 & 0.908 & 1.000 & 0.994 \\ \hline
\multicolumn{1}{|c|}{\cellcolor[HTML]{656565}4} & 0.945 & 0.901 & 0.994 & 1.000 \\ \hline
\end{tabular}
\end{center}

\noindent In conclusion, it appears that IDF weighting does not significantly matter for the results, nor does the distinction between $L_2$ and cosine metrics. However, both the choice of $L_1$ metric and the use of sampling and summarization make a difference.

\subsubsection{Stylometry}


\begin{center}
\begin{tabular}{cc|ccc|}
\cline{3-5}
&  & \multicolumn{3}{c|}{\textbf{length correction}} \\ \hline
\multicolumn{1}{|c|}{\textbf{metric}}        & \textbf{top words} & \multicolumn{1}{c|}{none}                      & \multicolumn{1}{c|}{sampling}                  & summ.                     \\ \hline
\multicolumn{1}{|c|}{cos}           & $50$      & \multicolumn{1}{c|}{\cellcolor[HTML]{EFEFEF}1} & \multicolumn{1}{c|}{\cellcolor[HTML]{EFEFEF}1} & \cellcolor[HTML]{9B9B9B}7 \\ \hline
\multicolumn{1}{|c|}{cos}           & $100$     & \multicolumn{1}{c|}{\cellcolor[HTML]{C0C0C0}2} & \multicolumn{1}{c|}{\cellcolor[HTML]{EFEFEF}1} & \cellcolor[HTML]{EFEFEF}5 \\ \hline
\multicolumn{1}{|c|}{cos}           & $200$     & \multicolumn{1}{c|}{\cellcolor[HTML]{EFEFEF}1} & \multicolumn{1}{c|}{\cellcolor[HTML]{EFEFEF}1} & \cellcolor[HTML]{EFEFEF}5 \\ \hline
\multicolumn{1}{|c|}{delta}         & any       & \multicolumn{1}{c|}{\cellcolor[HTML]{EFEFEF}1} & \multicolumn{1}{c|}{\cellcolor[HTML]{EFEFEF}1} & \cellcolor[HTML]{656565}4 \\ \hline
\multicolumn{1}{|c|}{argamon}       & any       & \multicolumn{1}{c|}{\cellcolor[HTML]{EFEFEF}1} & \multicolumn{1}{c|}{\cellcolor[HTML]{EFEFEF}1} & \cellcolor[HTML]{656565}4 \\ \hline
\multicolumn{1}{|c|}{eder}          & any       & \multicolumn{1}{c|}{\cellcolor[HTML]{EFEFEF}1} & \multicolumn{1}{c|}{\cellcolor[HTML]{EFEFEF}1} & \cellcolor[HTML]{656565}4 \\ \hline
\multicolumn{1}{|c|}{cross-entropy} & any       & \multicolumn{1}{c|}{\cellcolor[HTML]{EFEFEF}1} & \multicolumn{1}{c|}{\cellcolor[HTML]{EFEFEF}1} & \cellcolor[HTML]{656565}4 \\ \hline
\multicolumn{1}{|c|}{minmax}        & any       & \multicolumn{1}{c|}{\cellcolor[HTML]{EFEFEF}1} & \multicolumn{1}{c|}{\cellcolor[HTML]{EFEFEF}1} & \cellcolor[HTML]{656565}4 \\ \hline
\multicolumn{1}{|c|}{simple}        & any       & \multicolumn{1}{c|}{\cellcolor[HTML]{EFEFEF}1} & \multicolumn{1}{c|}{\cellcolor[HTML]{EFEFEF}1} & \cellcolor[HTML]{656565}4 \\ \hline
\multicolumn{1}{|c|}{cosine delta}  & any       & \multicolumn{1}{c|}{\cellcolor[HTML]{9B9B9B}3} & \multicolumn{1}{c|}{\cellcolor[HTML]{9B9B9B}3} & \cellcolor[HTML]{C0C0C0}6 \\ \hline
\end{tabular}
\end{center}

\begin{center}
\begin{tabular}{c|ccccccc|}
\cline{2-8}
& \multicolumn{7}{c|}{\textbf{Inter-cluster correlation matrix}} \\ \cline{2-8} 
& \multicolumn{1}{c|}{\cellcolor[HTML]{EFEFEF}1} & \multicolumn{1}{c|}{\cellcolor[HTML]{C0C0C0}2} & \multicolumn{1}{c|}{\cellcolor[HTML]{9B9B9B}3} & \multicolumn{1}{c|}{\cellcolor[HTML]{656565}4} & \multicolumn{1}{c|}{\cellcolor[HTML]{EFEFEF}5} & \multicolumn{1}{c|}{\cellcolor[HTML]{C0C0C0}6} & \cellcolor[HTML]{9B9B9B}7 \\ \hline
\multicolumn{1}{|c|}{\cellcolor[HTML]{EFEFEF}1} & \multicolumn{1}{c|}{$1.00$}                   & \multicolumn{1}{c|}{$.779$}                    & \multicolumn{1}{c|}{$.427$}                    & \multicolumn{1}{c|}{$.799$}                    & \multicolumn{1}{c|}{$.516$}                    & \multicolumn{1}{c|}{$.501$}                    & $.664$                    \\ \hline
\multicolumn{1}{|c|}{\cellcolor[HTML]{C0C0C0}2} & \multicolumn{1}{c|}{$.779$}                    & \multicolumn{1}{c|}{$1.00$}                   & \multicolumn{1}{c|}{$.340$}                    & \multicolumn{1}{c|}{$.602$}                    & \multicolumn{1}{c|}{$.672$}                    & \multicolumn{1}{c|}{$.296$}                    & $.535$                    \\ \hline
\multicolumn{1}{|c|}{\cellcolor[HTML]{9B9B9B}3} & \multicolumn{1}{c|}{$.427$}                    & \multicolumn{1}{c|}{$.340$}                    & \multicolumn{1}{c|}{$1.00$}                   & \multicolumn{1}{c|}{$.385$}                    & \multicolumn{1}{c|}{$.165$}                    & \multicolumn{1}{c|}{$.649$}                    & $.365$                    \\ \hline
\multicolumn{1}{|c|}{\cellcolor[HTML]{656565}4} & \multicolumn{1}{c|}{$.799$}                    & \multicolumn{1}{c|}{$.602$}                    & \multicolumn{1}{c|}{$.385$}                    & \multicolumn{1}{c|}{$1.00$}                   & \multicolumn{1}{c|}{$.687$}                    & \multicolumn{1}{c|}{$.572$}                    & $.779$                    \\ \hline
\multicolumn{1}{|c|}{\cellcolor[HTML]{EFEFEF}5} & \multicolumn{1}{c|}{$.516$}                    & \multicolumn{1}{c|}{$.672$}                    & \multicolumn{1}{c|}{$.165$}                    & \multicolumn{1}{c|}{$.687$}                    & \multicolumn{1}{c|}{$1.00$}                   & \multicolumn{1}{c|}{$.309$}                    & $.662$                    \\ \hline
\multicolumn{1}{|c|}{\cellcolor[HTML]{C0C0C0}6} & \multicolumn{1}{c|}{$.501$}                    & \multicolumn{1}{c|}{$.296$}                    & \multicolumn{1}{c|}{$.649$}                    & \multicolumn{1}{c|}{$.572$}                    & \multicolumn{1}{c|}{$.309$}                    & \multicolumn{1}{c|}{$1.00$}                   & $.556$                    \\ \hline
\multicolumn{1}{|c|}{\cellcolor[HTML]{9B9B9B}7} & \multicolumn{1}{c|}{$.664$}                    & \multicolumn{1}{c|}{$.535$}                    & \multicolumn{1}{c|}{$.365$}                    & \multicolumn{1}{c|}{$.779$}                    & \multicolumn{1}{c|}{$.662$}                    & \multicolumn{1}{c|}{$.556$}                    & $1.00$                   \\ \hline
\end{tabular}
\end{center}

The major distinctions can be observed between methods working on summarized and non-summarized texts, as well as between cosine-based metrics and norm-based metrics.

\subsubsection{Static Word Embeddings}


\begin{center}
\begin{tabular}{ccc|ccc|}
\cline{4-6}
&  &  & \multicolumn{3}{c|}{\textbf{length correction}} \\ \hline
\multicolumn{1}{|c|}{\textbf{method}} & \multicolumn{1}{c|}{\textbf{metric}} & \textbf{dim} & \multicolumn{1}{c|}{none}                      & \multicolumn{1}{c|}{sampling}                  & summ.             \\ \hline
\multicolumn{1}{|c|}{FastText}        & \multicolumn{1}{c|}{cos}             & any          & \multicolumn{1}{c|}{\cellcolor[HTML]{EFEFEF}1} & \multicolumn{1}{c|}{\cellcolor[HTML]{EFEFEF}1} & \cellcolor[HTML]{EFEFEF}1 \\ \hline
\multicolumn{1}{|c|}{FastText}        & \multicolumn{1}{c|}{wmd}             & any          & \multicolumn{1}{c|}{\cellcolor[HTML]{9B9B9B}2} & \multicolumn{1}{c|}{\cellcolor[HTML]{9B9B9B}2} & \cellcolor[HTML]{EFEFEF}1 \\ \hline
\multicolumn{1}{|c|}{GloVe}           & \multicolumn{1}{c|}{cos}             & any          & \multicolumn{1}{c|}{\cellcolor[HTML]{9B9B9B}2} & \multicolumn{1}{c|}{\cellcolor[HTML]{9B9B9B}2} & \cellcolor[HTML]{EFEFEF}1 \\ \hline
\multicolumn{1}{|c|}{GloVe}           & \multicolumn{1}{c|}{wmd}             & any          & \multicolumn{1}{c|}{\cellcolor[HTML]{9B9B9B}2} & \multicolumn{1}{c|}{\cellcolor[HTML]{9B9B9B}2} & \cellcolor[HTML]{EFEFEF}1 \\ \hline
\multicolumn{1}{|c|}{word2vec}        & \multicolumn{1}{c|}{cos}             & $100$, $300$ & \multicolumn{1}{c|}{\cellcolor[HTML]{9B9B9B}2} & \multicolumn{1}{c|}{\cellcolor[HTML]{EFEFEF}1} & \cellcolor[HTML]{EFEFEF}1 \\ \hline
\multicolumn{1}{|c|}{word2vec}        & \multicolumn{1}{c|}{cos}             & $500$, $800$ & \multicolumn{1}{c|}{\cellcolor[HTML]{9B9B9B}2} & \multicolumn{1}{c|}{\cellcolor[HTML]{9B9B9B}2} & \cellcolor[HTML]{EFEFEF}1 \\ \hline
\multicolumn{1}{|c|}{word2vec}        & \multicolumn{1}{c|}{wmd}             & any          & \multicolumn{1}{c|}{\cellcolor[HTML]{9B9B9B}2} & \multicolumn{1}{c|}{\cellcolor[HTML]{9B9B9B}2} & \cellcolor[HTML]{EFEFEF}1 \\ \hline
\end{tabular}
\end{center}

\noindent For FastText, but not for other models, we observe significant difference between cosine and wmd metrics. Summarization affects the results significantly, but sampling does not.

\subsubsection{Transformer Word Embeddings}


\begin{center}
\begin{tabular}{c|cccc|}
\cline{2-5}
& \multicolumn{4}{c|}{\textbf{pooling}} \\ \cline{2-5} 
& \multicolumn{2}{c|}{\textbf{mean}} & \multicolumn{2}{c|}{\textbf{max.}} \\ \hline
\multicolumn{1}{|c|}{\textbf{model}} & \multicolumn{1}{c|}{none} & \multicolumn{1}{c|}{summ.} & \multicolumn{1}{c|}{none} & summ.                 \\ \hline
\multicolumn{1}{|c|}{BART} & \multicolumn{1}{c|}{\cellcolor[HTML]{9B9B9B}3} & \multicolumn{1}{c|}{\cellcolor[HTML]{656565}4} & \multicolumn{1}{c|}{\cellcolor[HTML]{EFEFEF}1} & \cellcolor[HTML]{C0C0C0}2 \\ \hline
\multicolumn{1}{|c|}{RoBERTa-medium} & \multicolumn{1}{c|}{\cellcolor[HTML]{656565}4} & \multicolumn{1}{c|}{\cellcolor[HTML]{656565}4} & \multicolumn{1}{c|}{\cellcolor[HTML]{EFEFEF}1} & \cellcolor[HTML]{C0C0C0}2 \\ \hline
\multicolumn{1}{|c|}{RoBERTa-large}  & \multicolumn{1}{c|}{\cellcolor[HTML]{656565}4} & \multicolumn{1}{c|}{\cellcolor[HTML]{656565}4} & \multicolumn{1}{c|}{\cellcolor[HTML]{EFEFEF}1} & \cellcolor[HTML]{C0C0C0}2 \\ \hline
\multicolumn{1}{|c|}{GPT2-medium}    & \multicolumn{1}{c|}{\cellcolor[HTML]{656565}4} & \multicolumn{1}{c|}{\cellcolor[HTML]{656565}4} & \multicolumn{1}{c|}{\cellcolor[HTML]{EFEFEF}1} & \cellcolor[HTML]{C0C0C0}2 \\ \hline
\multicolumn{1}{|c|}{GPT2-xl}        & \multicolumn{1}{c|}{\cellcolor[HTML]{656565}4} & \multicolumn{1}{c|}{\cellcolor[HTML]{656565}4} & \multicolumn{1}{c|}{\cellcolor[HTML]{EFEFEF}1} & \cellcolor[HTML]{C0C0C0}2 \\ \hline
\multicolumn{1}{|c|}{LongFormer}     & \multicolumn{1}{c|}{\cellcolor[HTML]{656565}4} & \multicolumn{1}{c|}{\cellcolor[HTML]{656565}4} & \multicolumn{1}{c|}{\cellcolor[HTML]{EFEFEF}1} & \cellcolor[HTML]{C0C0C0}2 \\ \hline
\end{tabular}
\end{center}

\begin{center}
\begin{tabular}{c|cccc|}
\cline{2-5}
& \multicolumn{4}{c|}{\textbf{Inter-cluster correlation matrix}} \\ \cline{2-5} 
& \multicolumn{1}{c|}{\cellcolor[HTML]{EFEFEF}1} & \multicolumn{1}{c|}{\cellcolor[HTML]{C0C0C0}2} & \multicolumn{1}{c|}{\cellcolor[HTML]{9B9B9B}3} & \cellcolor[HTML]{656565}4 \\ \hline
\multicolumn{1}{|c|}{\cellcolor[HTML]{EFEFEF}1} & \multicolumn{1}{c|}{$1.000$}                   & \multicolumn{1}{c|}{$.799$}                    & \multicolumn{1}{c|}{$.469$}                    & $.614$                    \\ \hline
\multicolumn{1}{|c|}{\cellcolor[HTML]{C0C0C0}2} & \multicolumn{1}{c|}{$.799$}                    & \multicolumn{1}{c|}{$1.000$}                   & \multicolumn{1}{c|}{$.489$}                    & $.717$                    \\ \hline
\multicolumn{1}{|c|}{\cellcolor[HTML]{9B9B9B}3} & \multicolumn{1}{c|}{$.469$}                    & \multicolumn{1}{c|}{$.489$}                    & \multicolumn{1}{c|}{$1.000$}                   & $.607$                    \\ \hline
\multicolumn{1}{|c|}{\cellcolor[HTML]{656565}4} & \multicolumn{1}{c|}{$.614$}                    & \multicolumn{1}{c|}{$.717$}                    & \multicolumn{1}{c|}{$.607$}                    & $1.000$                   \\ \hline
\end{tabular}
\end{center}

\noindent As we can see, there are no significant differences between models, but the choice of pooling method matters. Sampling / summarization affects the results for max pooling, but not for mean pooling.

\subsubsection{Sentence Embeddings}


\begin{center}
\begin{tabular}{c|cccc|}
\cline{2-5}
& \multicolumn{4}{c|}{\textbf{pooling}} \\ \cline{2-5} 
& \multicolumn{2}{c|}{\textbf{mean}} & \multicolumn{2}{c|}{\textbf{max.}} \\ \hline
\multicolumn{1}{|c|}{\textbf{model}} & \multicolumn{1}{c|}{none} & \multicolumn{1}{c|}{summ.} & \multicolumn{1}{c|}{none} & summ. \\ \hline
\multicolumn{1}{|c|}{DistilRoBERTa}  & \multicolumn{1}{c|}{\cellcolor[HTML]{9B9B9B}2} & \multicolumn{1}{c|}{\cellcolor[HTML]{9B9B9B}2} & \multicolumn{1}{c|}{\cellcolor[HTML]{EFEFEF}1} & \cellcolor[HTML]{EFEFEF}1 \\ \hline
\multicolumn{1}{|c|}{MPNet2}         & \multicolumn{1}{c|}{\cellcolor[HTML]{9B9B9B}2} & \multicolumn{1}{c|}{\cellcolor[HTML]{9B9B9B}2} & \multicolumn{1}{c|}{\cellcolor[HTML]{EFEFEF}1} & \cellcolor[HTML]{EFEFEF}1 \\ \hline
\end{tabular}
\end{center}

\begin{center}
\begin{tabular}{c|cc|}
\cline{2-3}
& \multicolumn{2}{c|}{\textbf{Inter-cluster correlation matrix}} \\ \cline{2-3} 
& \multicolumn{1}{c|}{\cellcolor[HTML]{EFEFEF}\hspace{.9cm}1\hspace{.9cm}~} & \cellcolor[HTML]{9B9B9B}2 \\ \hline
\multicolumn{1}{|c|}{\cellcolor[HTML]{EFEFEF}1} & \multicolumn{1}{c|}{$1.000$}                   & $.755$                    \\ \hline
\multicolumn{1}{|c|}{\cellcolor[HTML]{9B9B9B}2} & \multicolumn{1}{c|}{$.755$}                    & $1.000$                   \\ \hline
\end{tabular}
\end{center}

Again we see correlation between models, but differences between pooling methods.

\subsection{Inter-Group Correlations}


Sentence embeddings and transformer-based word embeddings are strongly correlated, as is also the case with word frequency methods and stylometry. Word embedding methods are somewhat of an outlier, but closer to the latter. The relatively strong correlation between stylometry and transformer-based methods deserves a note.

\begin{center}
\begin{tabular}{c|c|c|c|c|c|}
\cline{2-6}
& \textbf{word} & \textbf{stylo-} & \textbf{word} & \textbf{trans-} & \textbf{sentence} \\
& \textbf{freq.} & \textbf{metry} & \textbf{embed.} & \textbf{formers} & \textbf{embed.} \\ \hline
\multicolumn{1}{|c|}{\textbf{word}} & \multirow{2}{*}{$1.000$} & \multirow{2}{*}{$.855$} & \multirow{2}{*}{$.684$} & \multirow{2}{*}{$.581$} & \multirow{2}{*}{$.522$}  \\
\multicolumn{1}{|c|}{\textbf{freq.}} & & & & & \\ \hline
\multicolumn{1}{|c|}{\textbf{stylo.}} & $.855$ & $1.000$ & $.522$ & $.855$ & $.793$ \\ \hline
\multicolumn{1}{|c|}{\textbf{word}} & \multirow{2}{*}{$.684$} & \multirow{2}{*}{$.522$}  & \multirow{2}{*}{$1.000$} & \multirow{2}{*}{$.533$} & \multirow{2}{*}{$.438$} \\
\multicolumn{1}{|c|}{\textbf{embed.}} & & & & & \\ \hline
\multicolumn{1}{|c|}{\textbf{trans.}} & $.581$ & $.855$ & $.533$ & $1.000$ & $.910$ \\ \hline
\multicolumn{1}{|c|}{\textbf{sentence}} & \multirow{2}{*}{$.522$} & \multirow{2}{*}{$.793$} & \multirow{2}{*}{$.438$} & \multirow{2}{*}{$.910$} & \multirow{2}{*}{$1.000$} \\
\multicolumn{1}{|c|}{\textbf{embed.}} & & & & & \\ \hline
\end{tabular}
\end{center}

\subsection{Benchmarks}

We note that most textual similarity results (with the exception of stylometric ones) perform similarly against expert assessments as document-based coding methods. Almost all methods perform poorly against behavioral benchmarks (voting, etc.), but this is more of a conceptual problem, as it affects MARPOR and other manifesto-based data sources as well.

\begin{center}
\begin{tabular}{|c|c|c|c|c|c|c|c|}
\hline
\textbf{group}  & \textbf{no.}                                            & \textbf{lrgen} & \textbf{lreco} & \textbf{galtan} & \textbf{ch2d} & \textbf{vdem}   & \textbf{rile}  \\ \hline
       & \cellcolor[HTML]{EFEFEF}1 & $.33$ & $.49$  & $.33$  & $.46$  & $.61$  & $.57$ \\ \cline{2-8} 
word   & \cellcolor[HTML]{C0C0C0}2 & $.46$ & $.59$  & $.46$  & $.59$  & $.84$  & $.55$ \\ \cline{2-8} 
       & \cellcolor[HTML]{9B9B9B}3 & $.48$ & $.48$  & $.48$  & $.56$  & $.74$  & $.52$ \\ \cline{2-8} 
\multirow{-2}{*}{freq.}  & \cellcolor[HTML]{656565}4 & $.55$ & $.53$  & $.55$  & $.64$  & $.85$  & $.58$ \\ \hline
       & \cellcolor[HTML]{EFEFEF}1 & $.11$ & $.16$  & $.11$  & $.06$  & $.09$  & $.21$ \\ \cline{2-8} 
       & \cellcolor[HTML]{C0C0C0}2 & $.08$ & $.21$  & $.08$  & $.11$  & $.06$  & $.28$ \\ \cline{2-8} 
stylo- & \cellcolor[HTML]{9B9B9B}3 & $.09$ & $.16$  & $.09$  & $.08$  & $.12$  & $.15$ \\ \cline{2-8} 
metry  & \cellcolor[HTML]{656565}4 & $.07$ & $.17$  & $.07$  & $.03$  & $.04$  & $.13$ \\ \cline{2-8} 
       & \cellcolor[HTML]{EFEFEF}5 & $.08$ & $.15$  & $.08$  & $.04$  & $.10$  & $.36$ \\ \cline{2-8} 
       & \cellcolor[HTML]{C0C0C0}6 & $.04$ & $.15$  & $.04$  & $.01$  & $-.04$ & $.14$ \\ \cline{2-8} 
       & \cellcolor[HTML]{9B9B9B}7 & $.08$ & $.14$  & $.08$  & $.04$  & $.06$  & $.31$ \\ \hline
word   & \cellcolor[HTML]{EFEFEF}1 & $.28$ & $.27$  & $.28$  & $.26$  & $.24$  & $.46$ \\ \cline{2-8} 
embed. & \cellcolor[HTML]{9B9B9B}2 & $.35$ & $.42$  & $.35$  & $.44$  & $.49$  & $.47$ \\ \hline
       & \cellcolor[HTML]{EFEFEF}1 & $.26$ & $.35$  & $.26$  & $.35$  & $.37$  & $.35$ \\ \cline{2-8} 
trans- & \cellcolor[HTML]{C0C0C0}2 & $.37$ & $.38$  & $.37$  & $.44$  & $.63$  & $.43$ \\ \cline{2-8} 
former & \cellcolor[HTML]{9B9B9B}3 & $.46$ & $.48$  & $.46$  & $.51$  & $.69$  & $.48$ \\ \cline{2-8} 
       & \cellcolor[HTML]{656565}4 & $.45$ & $.44$  & $.45$  & $.48$  & $.61$  & $.39$ \\ \hline
sent.  & \cellcolor[HTML]{EFEFEF}1 & $.21$ & $.31$  & $.21$  & $.31$  & $.35$  & $.29$ \\ \cline{2-8} 
embed. & \cellcolor[HTML]{9B9B9B}2 & $.38$ & $.41$  & $.38$  & $.43$  & $.45$  & $.37$ \\ \hline
\end{tabular}
\end{center}

\begin{center}
\begin{tabular}{|c|c|c|c|c|c|c|}
\hline
\textbf{group}  & \textbf{no.}                                            & \textbf{marpor} & \textbf{vote}   & \textbf{coal}   & \textbf{cand}   & \textbf{elec}   \\ \hline
       & \cellcolor[HTML]{EFEFEF}1 & $.25$  & $.00$  & -$.13$ & $.02$  & $.02$  \\ \cline{2-7} 
word   & \cellcolor[HTML]{C0C0C0}2 & $.44$  & $.36$  & $.28$  & $.03$  & $.05$  \\ \cline{2-7} 
       & \cellcolor[HTML]{9B9B9B}3 & $.10$  & -$.01$ & -$.05$ & $.05$  & -$.04$ \\ \cline{2-7} 
\multirow{-2}{*}{freq.}  & \cellcolor[HTML]{656565}4 & $.24$  & $.19$  & $.30$  & $.13$  & $.31$  \\ \hline
       & \cellcolor[HTML]{EFEFEF}1 & $.23$  & $.05$  & -$.01$ & $.14$  & -$.07$ \\ \cline{2-7} 
       & \cellcolor[HTML]{C0C0C0}2 & $.18$  & -$.08$ & -$.09$ & $.13$  & -$.09$ \\ \cline{2-7} 
stylo- & \cellcolor[HTML]{9B9B9B}3 & $.09$  & $.21$  & $.14$  & $.12$  & -$.05$ \\ \cline{2-7} 
metry  & \cellcolor[HTML]{656565}4 & $.20$  & $.01$  & -$.02$ & $.14$  & -$.13$ \\ \cline{2-7} 
       & \cellcolor[HTML]{EFEFEF}5 & $.12$  & $.10$  & $.13$  & $.19$  & $.15$  \\ \cline{2-7} 
       & \cellcolor[HTML]{C0C0C0}6 & -$.01$ & $.09$  & $.03$  & $.15$  & -$.11$ \\ \cline{2-7} 
       & \cellcolor[HTML]{9B9B9B}7 & $.10$  & -$.05$ & -$.12$ & $.16$  & -$.15$ \\ \hline
word   & \cellcolor[HTML]{EFEFEF}1 & -$.02$ & -$.02$ & -$.07$ & $.14$  & -$.03$ \\ \cline{2-7} 
embed. & \cellcolor[HTML]{9B9B9B}2 & $.07$  & -$.06$ & -$.08$ & $.01$  & $.00$  \\ \hline
       & \cellcolor[HTML]{EFEFEF}1 & $.21$  & -$.10$ & -$.11$ & $.03$  & $.01$  \\ \cline{2-7} 
trans- & \cellcolor[HTML]{C0C0C0}2 & $.06$  & $.00$  & -$.02$ & $.03$  & -$.02$ \\ \cline{2-7} 
former & \cellcolor[HTML]{9B9B9B}3 & -$.04$ & $.06$  & $.01$  & $.06$  & $.01$  \\ \cline{2-7} 
       & \cellcolor[HTML]{656565}4 & $.16$  & $.05$  & $.06$  & $.09$  & -$.05$ \\ \hline
sent.  & \cellcolor[HTML]{EFEFEF}1 & $.21$  & -$.09$ & -$.07$ & -$.03$ & -$.01$ \\ \cline{2-7} 
embed. & \cellcolor[HTML]{9B9B9B}2 & $.31$  & -$.04$ & -$.08$ & -$.01$ & -$.05$ \\ \hline
\end{tabular}
\end{center}

\section{Future Work}

Future work will focus on testing \emph{additional methods}, including LDA-based topic models with scaling and methods based on topic matching; exploring the potential of combining textual similarity methods with machine translation algorithms to obtain \emph{inter-language comparability}; and aggregating textual similarity measures to algorithmically \emph{recover party positions} in the policy space.

{
\footnotesize
\bibliography{partysim}
}

\end{document}